\documentclass[
]{ceurart}

\usepackage{listings}
\begin{document}

\copyrightyear{2022}
\copyrightclause{Copyright for this paper by its authors.
  Use permitted under Creative Commons License Attribution 4.0
  International (CC BY 4.0).}

\conference{ISWC'22: The 21st International Semantic Web Conference,
  October 23--27, 2022, Hangzhou, China}

\title{SignalKG: Towards Reasoning about the Underlying Causes of Sensor Observations}

\author{Anj Simmons}[%
orcid=0000-0001-8402-2853,
email=a.simmons@deakin.edu.au
]

\author{Rajesh Vasa}[%
orcid=0000-0003-4805-1467,
email=rajesh.vasa@deakin.edu.au
]

\author{Antonio Giardina}[%
orcid=0000-0003-1047-6339,
email=antonio.giardina@deakin.edu.au
]

\address{Applied Artificial Intelligence Institute, Deakin University, Geelong, Australia}


\begin{abstract}
  This paper demonstrates our vision for knowledge graphs that assist machines to reason about the cause of signals observed by sensors. We show how the approach allows for constructing smarter surveillance systems that reason about the most likely cause (e.g., an attacker breaking a window) of a signal rather than acting directly on the received signal without consideration for how it was produced.
\end{abstract}

\begin{keywords}
  knowledge graph \sep
  ontology \sep
  sensor \sep
  surveillance
\end{keywords}

\maketitle

\section{Introduction}

Standards such as the Semantic Sensor Network (SSN/SOSA) ontology \cite{ArminHaller2018} allow capturing the semantics of sensor observations, and emerging standards for smart buildings \cite{Hammar2019,rasmussen2021bot} and smart cities allow capturing the semantics of the environments in which sensors operate. However, reasoning about the underlying cause of sensor observations requires not only knowledge of the sensors and their environment, but also an understanding of the signals they detect and the possible causes of these signals. For example, inferring that the sound of breaking glass may be due to a broken window requires knowledge of the fact that glass windows produce a distinct sound when broken and that this sound propagates as sound waves through the air to a sensor such as a human ear or microphone.
This paper proposes a signal knowledge graph (SignalKG) to support machines to reason about the underlying cause of sensor observations. Sensors and their environment are represented using existing standards, and then linked to SignalKG. To reason about the cause of sensor observations, we automatically generate a Bayesian network based on information in the knowledge graph, and use this to infer the posterior probability of causes given the sensor data.

\begin{figure*}[tpb]
    \centering
    \includegraphics[width=\linewidth]{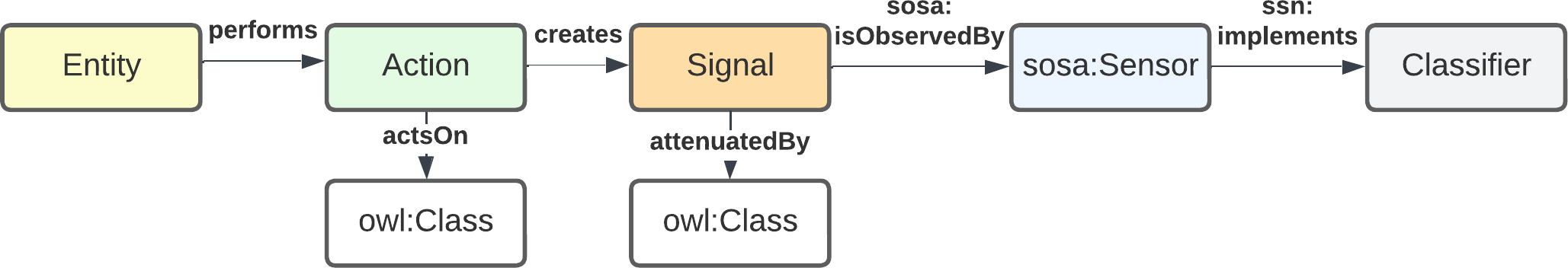}
    \caption{High level overview of SignalKG ontology}
    \label{fig:signal-kg-full}
\end{figure*}


\autoref{fig:signal-kg-full} presents an ontology describing the high level concepts.
A category of \textit{entities} (e.g., humans) perform \textit{actions} (e.g., walking) that act on a type of object or place (e.g., in hallways), which in turn create a type of \textit{signal} (e.g., the sound of footsteps). A \textit{sensor} observes a particular type of signal, which usually reduces in strength with distance and can be distorted by surrounding objects (e.g., a wall between a sound source and the receiver attenuates the sound signal). The sensor may implement a \textit{classifier} to detect the presence of the signal (e.g., an acoustic detector may make use of a binary machine learning classifier to detect if the sound of footsteps is present or not).
A formal RDF/OWL specification of the SignalKG ontology is available online\footnote{SignalKG ontology: \url{https://signalkg.visualmodel.org/skg}} as well as an \textit{interactive demonstration}\footnote{Interactive demonstration available at: \url{https://signalkg.visualmodel.org}} of how SignalKG can be applied and used to reason about the underlying causes of sensor observations.

\section{Related Work}
\label{sec:related-work}

Past research has utilised ontologies for the purpose of threat and situation awareness using description logic \cite{Roy2012} and rule based reasoning \cite{Pai2017}. However, these approaches assume that threats can be classified into classes according to deterministic rules, whereas in reality threats may be probabilistic in nature and there may be multiple possible explanations for a given set of observations. To overcome this limitation, reasoning methods have been proposed that combine deterministic rules with a Bayesian network for reasoning probabilistically \cite{Yao2022}. However, the structure of the Bayesian network and associated probabilities need to be manually specified. In contrast, we seek to express this knowledge in a reusable and extensible form. There have been attempts to extend OWL to support representing/reasoning about uncertainty \cite{Carvalho2017}. However, general approaches do not directly specify concepts for reasoning about sensor signals.

\section{Demonstration}

\begin{figure}[tpb]
    \centering
    \includegraphics[width=\linewidth]{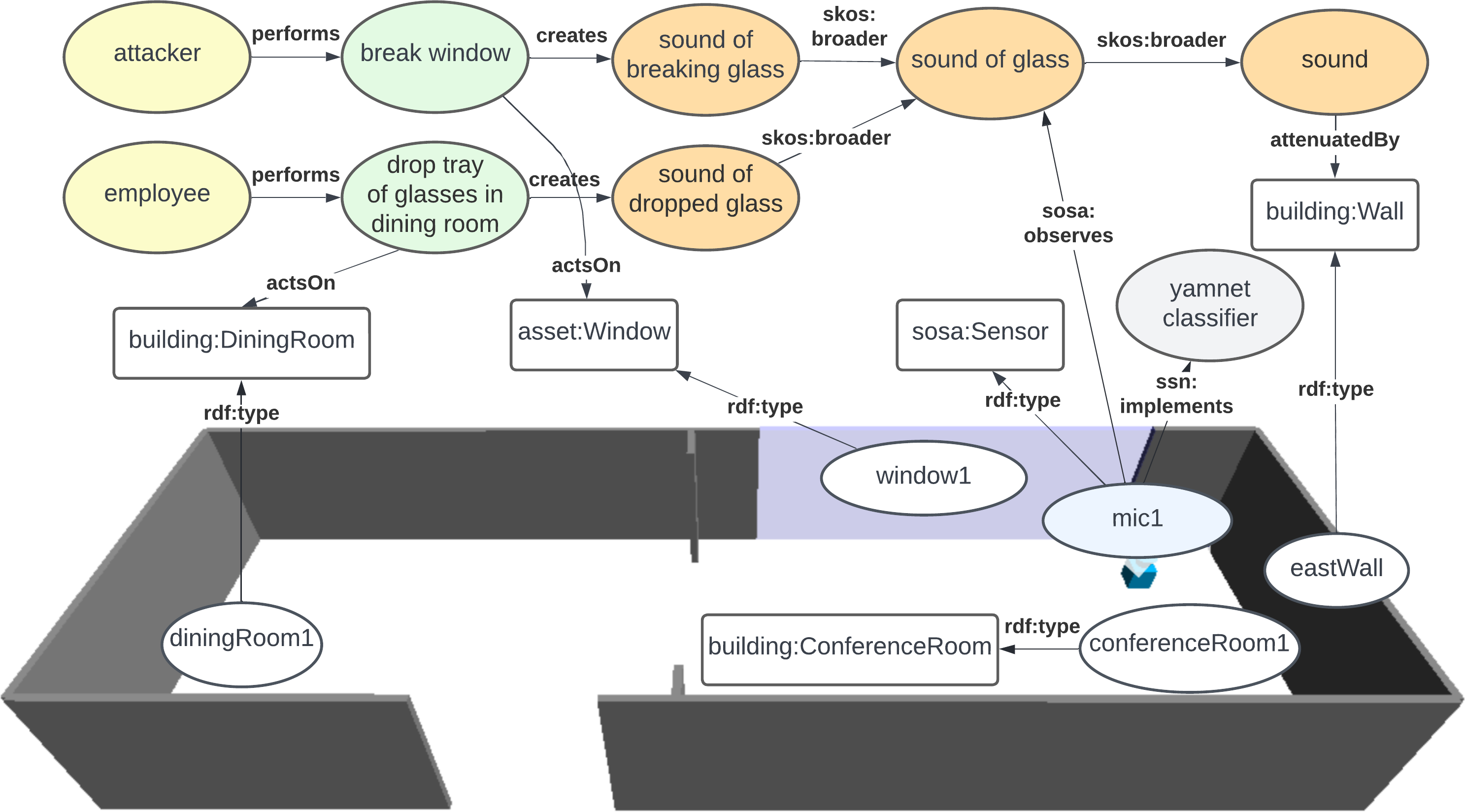}
    \caption{Knowledge graph of audio signals constructed using SignalKG ontology showing link to building rooms/assets and sensors}
    \label{fig:sound-model}
\end{figure}

\autoref{fig:sound-model} demonstrates how concepts in the SignalKG ontology can be applied to construct a knowledge graph of audio signals. Attackers and employees are entities that are capable of producing an action, such as breaking a window or dropping a tray of glasses in a dining room. Actions create signals, in this case, the sound of breaking glass or the sound of dropped glass. The building asset/room type on/in which an action can occur is represented using the RealEstateCore ontology \cite{Hammar2019}. To group similar signals together, we use the Simple Knowledge Organization System (SKOS) \cite{miles2009skos} `broader' property to represent a signal hierarchy, for example, the sound of breaking glass is similar to the sound of dropped glass and thus are grouped under the same category. The knowledge graph also includes information about how signals propagate, for example, that sound intensity reduces with distance (according to an inverse square law) and is attenuated by walls. To model the case in which signals are classified on the sensing device, we allow for describing different classification models, such as YAMNet, an audio classifier that recognises 521 classes of sounds (e.g., glass).

\begin{figure}[tpb]
    \centering
    \includegraphics[width=0.4\linewidth]{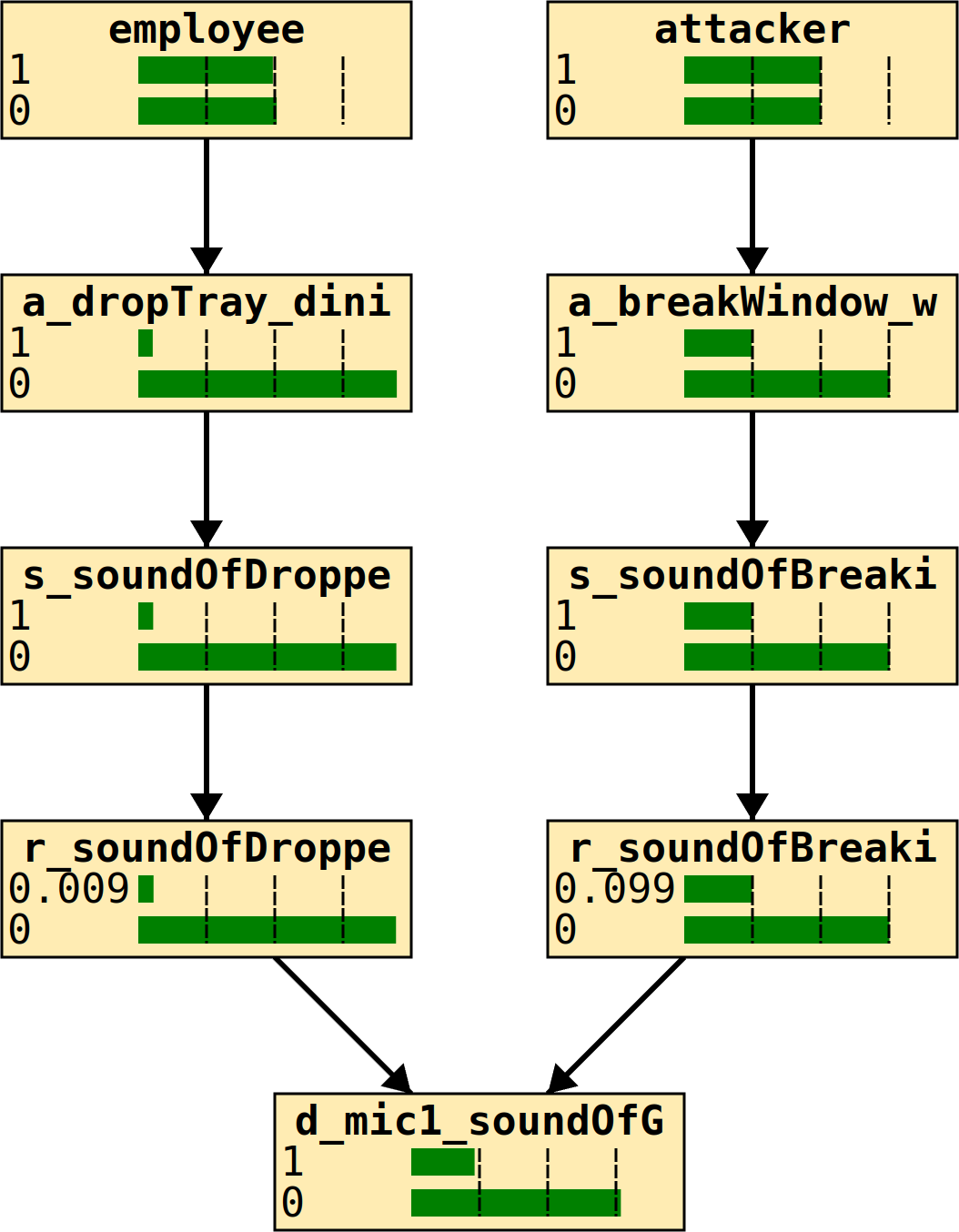}
    \hspace{2cm}
    \includegraphics[width=0.4\linewidth]{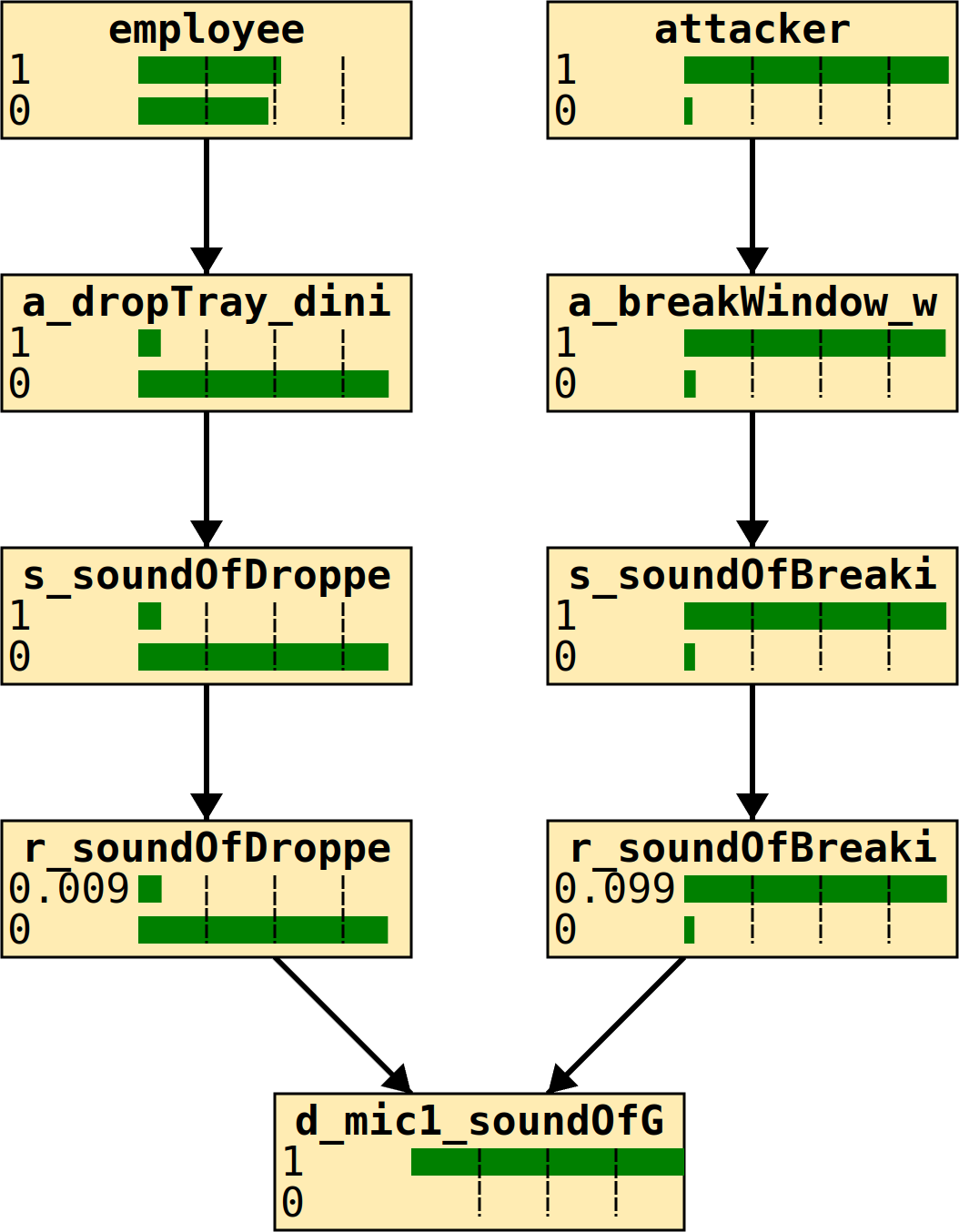}
    \caption{Bayesian network before (left) and after (right) conditioning on sensor observations. Green bars indicate the probability of each node value. (Key: a=action, s=signal emitted due to action, r=received signal strength at location of sensor, d=detected signal after classification). Visualisation created with jsbayes-viz \cite{vang_jsbayesviz_2016}.}
    \label{fig:sound-model-bayes}
\end{figure}

Sensor observations can be linked to the signal knowledge graph via the property (signal) that the sensor observes. Our interactive demonstration includes a simulator to generate sample sensor observations (represented using the SSN/SOSA ontology) for hypothetical scenarios that could occur. The goal is to infer what took place given only the sensor observations, knowledge of the building and sensor placement, and our understanding of possible underlying causes of observed signals (specified using SignalKG).

To support reasoning probabilistically about causes of sensor observations, the knowledge graph also includes probabilities, such as the prior probability that an entity will be present, and the probability that an entity (if present) will perform an action. As our goal is to infer a cause given observations, it lends itself to Bayesian reasoning. Rather than manually specifying a Bayesian network for a particular scenario, we automatically generate one based on the information in the knowledge graph. Encoding all the information needed to reason about signals in the knowledge graph itself helps facilite reuse and extension.

Nodes in the Bayesian network are generated for each entity, action at a location, signal emitted/received, and sensor. While our simple example results in a 1:1 mapping (shown in \autoref{fig:sound-model-bayes}), in more complex scenarios there may be vastly more nodes due to all the possible permutations of entity, action, location, signal and sensor. If multiple types of signals are present (audio, vision, social, etc.) then these all appear as part of the same generated Bayesian network, e.g., an attacker breaking a window will create both audio (sound of breaking glass) and vision (suspicious behaviour in a video feed) based signals. Prior probabilities for entities and actions need to be specified in the knowledge graph. The probability that a signal emitted by an action will be detected by a sensor is calculated based on the distance from the location of the action to the location of the sensor, how the signal intensity reduces with distance (e.g., the knowledge graph may specify an inverse square law for sound signals), any barriers between the source and the sensor that may attenuate/block the signal (e.g., the knowledge graph may specify sound is attenuated by walls), and the sensitivity of the classifier used by the sensor to detect presence of a signal.
 
Once we have generated a Bayesian network, we can condition it on sensor observations to infer the posterior probability of the underlying cause. For the demonstration, we estimate the posterior probability via likelihood weight sampling, implemented by \cite{vang_jsbayesviz_2016}, drawing 20,000 samples (the number of samples to draw is a trade-off between accuracy and computation time).
In the example, prior to conditioning on observations, there is a 50\% chance of an attacker being present. After conditioning on the observation that the microphone has detected the sound of glass, the posterior probability of an attacker increases to 97\%.

\section{Next Steps}

Even for the simple example of detecting building intrusions, the space of possible causes is large (e.g., an attacker could impersonate an employee then ask someone to let them in, or suspicious sounds could be due to a movie playing in the background). Furthermore, signals are not independent (as assumed by our preliminary prototype), but rather occur in sequences (e.g., the sound of footsteps, followed by a weapon detected in video footage, followed by a scream) that could help more reliably distinguish between possible causes. Also, more realistic models of signal propagation are needed, which may require continuous probability distributions rather than a conditional probability table over a discrete set of values as in our example Bayesian network. To support efficient reasoning about these more complex scenarios, we plan to explore generation of probabilistic programs in place of the discrete Bayesian networks used in this paper.

A practical barrier to uptake of our approach is the need to specify prior probabilities for each action that an entity can perform. Ordinary behaviour could potentially be learned from data. However, modelling prior probabilities of actions intruders perform is more difficult, as attacks are rare events (limited data to learn from) and an adversary will adjust their actions to avoid detection. In future work, we plan to include the goals of the intruder as part of the knowledge graph, then use a game theoretic approach to determine probable actions they will take rather than manually specifying prior probabilities for each action.

\begin{acknowledgments}
This research was funded by National Intelligence Postdoctoral Grant NIPG-2021-006.
\end{acknowledgments}

\bibliography{refs}


\end{document}